\documentclass{article} 
\usepackage[final]{colm2025_conference}

\usepackage{microtype}
\usepackage{hyperref}
\usepackage{url}
\usepackage{booktabs}
\usepackage{pythonhighlight}

\usepackage{lineno}

\usepackage{times}
\usepackage{latexsym}
\usepackage{amsmath}
\usepackage{blindtext}
\usepackage{graphicx}
\usepackage{enumerate}
\usepackage{caption}
\usepackage{subcaption}
\usepackage{multicol}
\usepackage{mathptmx}
\usepackage{amsmath}
\usepackage{amssymb}
\usepackage{nameref}
\usepackage{graphicx}
\usepackage{hyperref}

\usepackage[T1]{fontenc}

\usepackage[utf8]{inputenc}

\usepackage{microtype}

\usepackage{inconsolata}

\usepackage{graphicx}

\definecolor{darkblue}{rgb}{0, 0, 0.5}

\title{Exploring Neural Ordinary Differential Equations as Interpretable Healthcare classifiers}

\author{Shi Li \\
  Department of Computer Science \\
  Columbia University \\
  \texttt{shili081100@columbia.edu}
}

\begin{document}

\ifcolmsubmission
\linenumbers
\fi

\maketitle

\begin{abstract}
Deep Learning has emerged as one of the most significant innovations in machine learning. However, a notable limitation of this field lies in the ``black box" decision-making processes, which have led to skepticism within groups like healthcare and scientific communities regarding its applicability. In response, this study introduces a interpretable approach using Neural Ordinary Differential Equations (NODEs), a category of neural network models that exploit the dynamics of differential equations for representation learning. Leveraging their foundation in differential equations, we illustrate the capability of these models to continuously process textual data, marking the first such model of its kind, and thereby proposing a promising direction for future research in this domain. The primary objective of this research is to propose a novel architecture for groups like healthcare that require the predictive capabilities of deep learning while emphasizing the importance of model transparency demonstrated in NODEs.
\end{abstract}

\section{Introduction}

Deep learning has been recognized as a major innovation in machine learning \citep{lecun2015deep}, transforming the analysis of data and the derivation of insights across various domains, including computer vision \citep{voulodimos2018deep} and natural language processing \citep{deng2018deep, ono2024text}. Its capacity to learn hierarchical data representations renders it an influential tool, facilitating the automation of tasks previously deemed too complex or unattainable. Nonetheless, despite its significant contributions, deep learning presents notable challenges \citep{thompson2020computational}. Among these is the issue of interpretability (\cite{zhang2021survey}, \cite{chakraborty2017interpretability}, \cite{zhang2018visual}); particularly, models with intricate architectures tend to function as ``black boxes". The absence of transparency raises substantial concerns in critical areas such as healthcare (\citep{miotto2018deep}, \citep{razzak2018deep}) and finance \citep{heaton2016deep}, where the clarity of operations is imperative for ensuring trust, meeting ethical standards, and adhering to regulatory requirements \citep{wu2021medical}.

In contrast with recent emerging NLP techniques (e.g., GPT, BERT, etc.), our focus centers on a specific category of neural networks referred to as Neural Ordinary Differential Equations (NODEs) \citep{chen2018neural}, renowned for their exceptional performance across a multitude of data formats. Here, we present an extensive exploration of their utility within the textual modality, an area that has not been extensively explored. This paper advocates for the efficacy of NODEs in handling textual data, particularly in fields where transparency and model interpretability are crucial. 




\section{Related Works}

\subsection{Neural ODEs}

Neural Ordinary Differential Equations (NODEs), introduced in 2018 by \citep{chen2018neural}, have garnered widespread interest across various disciplines. These models offer a distinct advantage by continuously updating their internal state, facilitating seamless integration with time-series data and complex dynamic systems \citep{kidger2022neural}. Demonstrating exceptional performance, NODEs have found applications in tasks such as time-series forecasting (\cite{kidger2020neural}, \cite{jin2022multivariate}), dynamic system modeling (\cite{linot2023stabilized}, \cite{alvarez2020dynode}), and function approximation \citep{ruiz2023neural}. Furthermore, their interpretable nature and capability to capture continuous dynamics make them well-suited for healthcare applications, including medical image analysis and text-based outcome prediction. Although there has been limited exploration of time series analysis within Neural ODEs and healthcare (\cite{qian2021integrating}, \cite{gwak2020neural}), their potential in these other domains remains promising.

\subsection{Interpretability}

Interpretable deep learning refers to the development of machine learning models that produce results in a transparent and understandable manner (\cite{li2022interpretable}, \cite{chen2019looks}). Unlike traditional deep learning models, which often function as "black boxes" with opaque decision-making processes, interpretable deep learning methods aim to provide insights into how and why a model arrives at a particular outcome. By incorporating features such as interpretable attention mechanisms \citep{choi2016retain}, explainable embeddings \citep{subramanian2018spine}, and model-agnostic interpretability techniques, interpretable deep learning enhances trust and confidence in the predictions generated by these models. 







\section{Methods}
\subsection{A Primer on Neural ODEs}

\begin{table*}
  \centering
  \begin{tabular}{llllll}
    \hline
    \textbf{Model} & \textbf{Interpretable?} & \textbf{Accuracy} & \textbf{F1} & \textbf{AUROC} & \textbf{}\\
    \hline
    Logistic Regression &  yes & 0.913 & 0.914 & 0.963\\
    LGBM     &  yes& 0.928 & 0.929 & 0.980  \\
    LSTM & no & 0.932 &  0.933 & 0.929 \\
    BERT & no & 0.940 & 0.943 & 0.948 \\
    Neural ODE &  yes& 0.930& 0.932& 0.937\\
    
    \hline
  \end{tabular}
  \caption{Benchmark of text classification.}
  \label{tab16}
\end{table*}

Neural Ordinary Differential Equations (NODEs) fundamentally transform machine learning by treating the learning of continuous dynamics as solving an ordinary differential equation (ODE) \citep{kidger2022neural}. Central to NODEs is the concept that the derivative of the hidden state of a neural network with respect to time, denoted as \(\frac{d\mathbf{h}(t)}{dt}\), is a function \(f\) parameterized by a neural network. Formally, this relationship is expressed as \(\frac{d\mathbf{h}(t)}{dt} = f(\mathbf{h}(t), t; \theta)\), where \(\mathbf{h}(t)\) represents the hidden state at time \(t\), and \(\theta\) denotes the neural network parameters.

Given an initial condition \(\mathbf{h}(t_0) = \mathbf{h}_0\), the solution to this ODE, \(\mathbf{h}(t)\), at any time \(t\) is obtained by solving the integral \(\mathbf{h}(t) = \mathbf{h}_0 + \int_{t_0}^{t} f(\mathbf{h}(s), s; \theta) ds\). This integral captures the evolution of the network's hidden state over time, allowing for modeling continuous-time dynamical systems without fixed-step discretization, as in traditional recurrent neural networks (RNNs).

To train a Neural ODE, the adjoint method \citep{zhuang2020adaptive} is commonly employed, a backpropagation technique that efficiently computes gradients without storing intermediate states. The adjoint state, denoted as \(\mathbf{a}(t) = \frac{\partial L}{\partial \mathbf{h}(t)}\), where \(L\) is the loss function, evolves according to the ODE \(-\frac{d\mathbf{a}(t)}{dt} = \mathbf{a}(t)^T \frac{\partial f(\mathbf{h}(t), t; \theta)}{\partial \mathbf{h}(t)}\), with the initial condition set at the end of the interval and integration performed backward in time. The gradient of the loss \(L\) with respect to the parameters \(\theta\) is then obtained by integrating another term, \(\frac{\partial L}{\partial \theta} = -\int_{t}^{t_0} \mathbf{a}(t)^T \frac{\partial f(\mathbf{h}(t), t; \theta)}{\partial \theta} dt\), over the same interval in the opposite direction. A basic example of a NODE implementation is provided in the appendix (Section \ref{arc}).

\subsection{Current Gaps in Text Classification?}

Deep learning-based models (RNNs, LSTMs, BERT, etc.) for text classification face significant challenges in terms of interpretability. These models are often perceived as black boxes due to their complex architectures and opaque decision-making processes. Understanding which features or words influence the model's predictions is difficult, limiting the ability to trust and interpret the model's outputs. Techniques such as attention mechanisms provide some insights into the model's reasoning, but comprehensive explanations are often lacking. Furthermore, issues such as dataset biases and model errors can be challenging to diagnose and address without transparent interpretability tools. 

\subsection{How NODEs can model Text}

In the context of our work, modeling text for classification using NODEs offers several advantages. Firstly, NODEs represent text as a continuous dynamic system, enabling a more natural representation compared to traditional approaches. They excel in capturing long-range dependencies in text, which can be challenging for RNNs. NODEs also provide flexibility in handling sequences of varying lengths, eliminating the need for padding or truncation. Moreover, their interpretable dynamics allow for a deeper understanding of how textual features evolve over time, crucial for tasks where interpretability is paramount. Additionally, NODEs can be trained efficiently using the adjoint method, making them scalable for training on large text corpora. 

\section{Results}
\subsection{Case Study: Hospital Outcome Prediction}

Deep learning has garnered considerable attention in healthcare, spanning from predicting medical outcomes \citep{lee2024multimodal, jenkins2018dynamic, lee2024enhancing} to diagnosing diseases \citep{liu2018deep, cheng2016risk}, operationalization of medical tasks \citep{lee2024can} offering potential assistance across various clinical operations. However, lingering concerns about model interpretability persists \citep{stiglic2020interpretability, wu2010prediction}. Transparency is particularly crucial, especially with the emergence of Generative AI techniques, given their propensity for producing hallucinations, which can be harmful in healthcare contexts \citep{lee2024large}.

Therefore, in this case study, we employ NODEs for text classification, marking a pioneering application of this model type for such an objective. Framing this as a classification problem, our model adeptly learns the sequencing of text to discern the ``dynamics'' necessary for making predictions. For our study, we utilize classification objectives outlined in \citep{lee2024multimodal} to predict Emergency Department admissions, with additional results on extended tasks provided in the appendix using text serialized from MIMIC-IV ED database \citep{Johnson2023MIMICIVED}. Before inputting the serialized text into the NODE, we preprocess it using TF-IDF. 

To evaluate the effectiveness of the NODE, we compare its performance against two interpretable models, namely Logistic Regression and Light Gradient Boosted Machines (LGBM), as well as two non-interpretable models tailored for text processing, namely Long Short-Term Memory (LSTMs) and BERT. Our assessment focuses on the task of outcome prediction, and the results are presented in Table \ref{tab11}. This approach sheds light on the potential of NODEs in enhancing text classification tasks and underscores the importance of evaluating model interpretability alongside predictive performance.

\begin{table}
  \centering
  \begin{tabular}{ll}
    \hline
    \textbf{Admit Words} & \textbf{Home Words} \\
    \hline
    old & patient \\ 
    complaint & recieved\\
    diagnostic & 17 \\
    ambulance & walk-in\\
    insurance & medications\\
    
    \hline
  \end{tabular}
  \caption{Top 5 words for each basing prediction}
  \label{tab12}
\end{table}

\begin{table*}
  \centering
  \begin{tabular}{lll}
    \hline
    \textbf{Model} & \textbf{Interpretable?} & \textbf{Balanced Accuracy} \\
    \hline
    SVM &  somewhat & 0.611 \\
    CNN & somewhat & 0.781   \\
    VGG & no & 0.729 \\
    Neural ODE &  yes & 0.689\\
    
    \hline
  \end{tabular}
  \caption{Benchmark of image classification.}
  \label{tab15}
\end{table*}

\subsubsection{Feature Importance with Saliency Maps}

One advantage of the Neural ODE model over other attention-based approaches, such as BERT and LSTMs, is its ability to provide interpretable feature importance which can be directly mapped to words within the TF-IDF matrix via saliency maps. Saliency maps \citep{adebayo2018sanity} offer an interpretable method to comprehend the decision-making process of machine learning models and reveal which words or sequences contributed most to the classification decision, thereby offering insights into the model's reasoning. We therefore detail the saliency map from our hospital admission problem (Table \ref{tab12}), where specific keywords impacted the overall decision assigned by the model. This interpretability stands in contrast to the complex, hidden layers of BERT and recent LLMs, where understanding the precise influence of each token element on the output requires intricate analysis.

\subsubsection{Vector Fields explain Decision Making}

\begin{figure}[ht]
    \centering
    \includegraphics[width=0.8\columnwidth]{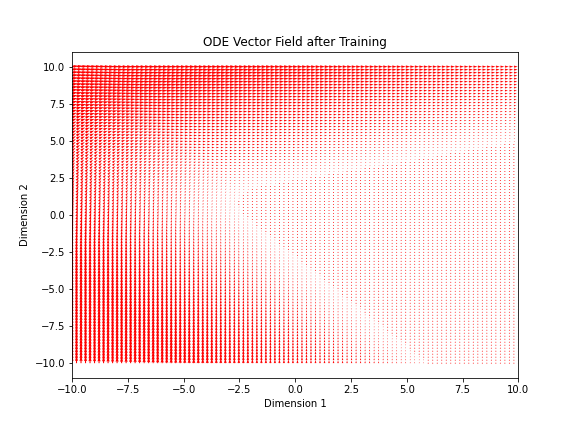}
    \caption{Visualizing the ODE Vector fields for a simple binary classification}
    \label{vf1}
\end{figure}

Another method of visualization utilizes ODE vector fields within the NODEs architecture, enhancing the models' interpretability (Figure \ref{vf1}). This approach enables intuitive visual demonstrations of high-dimensional data classification on a 2D plane. By modeling the dynamics of the neural network as an ODE, NODEs provide a vivid visualization of the hidden state's evolution over time in reaction to input data. This evolution, captured in the vector field, visually represents the direction and magnitude of change at various points in the input space. By analyzing the vector field's flow, insights into the model's processing and classification of input data at attractors can be gained. Visualizing these trajectories allows for an intuitive understanding of how the model differentiates between data classes and makes classification decisions based on the NODE's encoded dynamics.

\subsection{Case Study: Predicting the Stage of Alzheimer's}

Another potential application of Neural Ordinary Differential Equations (ODEs) in healthcare is in medical imaging, where they can adopt a similar approach to the text classification example discussed earlier. Instead of analyzing sequences of text, Neural ODEs can be applied to sequences of pixels representing medical images. Image classification with Neural ODEs involves treating the pixels of an image as a sequence, similar to words in a sentence, and learning the dynamics of these pixels to distinguish unique classes. Mathematically, let \( X(t) \) represent the state of the pixel sequence at time \( t \), and \( f(X(t), t; \theta) \) denote the continuous dynamics function parameterized by \( \theta \). The model continuously updates its internal state according to the differential equation:\( \frac{dX}{dt} = f(X(t), t; \theta) \)as it processes the sequence of pixels. This equation captures the complex spatial relationships within the image over time. By integrating the principles of differential equations into the learning process, Neural ODEs offer a unique advantage in image classification tasks by leveraging the continuous and interpretable nature of their dynamics to make predictions.


\section{Discussion}

\subsection{Continuous Modeling is Competitive to Discrete Modeling}

Our results reveals that the NODEs performs comparably in accuracy to alternative models. However, continuous modeling approaches present distinct advantages over discrete methodologies in text classification tasks. Specifically, continuous modeling adeptly captures the evolving dynamics of text, facilitating the extraction of expressive and flexible features. This capability is particularly beneficial in contexts where understanding the nuanced aspects of language is essential for precise classification. Furthermore, unlike traditional neural networks, NODEs are not susceptible to the vanishing or exploding gradient problems (Section \ref{grad}), potentially ensuring more stable and effective training outcomes. By harnessing the continuous-time framework of the ODE, NODEs are adept at both interpolating and extrapolating text representations beyond the confines of observed data, significantly improving their generalization to new inputs.

\subsection{Shaving Accuracy for Interpretability}

One observation we made from our analysis was that there exists a trade-off between interpretability and accuracy with the NODEs architecture. Consequently, NODEs might compromise a degree of accuracy compared to more intricate deep learning architectures in NLP that place a higher emphasis on performance over interpretability, as evidenced by our findings. Although LSTMs and BERT models may be favored for their high accuracy, the value of NODEs lies in their interpretability. 

\subsection{Performance versus Memory Tradeoff}

Compared to traditional discrete models, NODEs offer an additonal different trade-off in terms of time complexity and memory requirements. While the time complexity of training NODEs scales with the number of data points, the ODE solver steps, and the dimension of the state, it can be more efficient than the sequence-level computations required by full attention mechanisms, especially for long input sequences. Additionally, NODEs have the potential to be more memory-efficient, as they do not need to store the entire input sequence during the forward pass, but rather integrate the dynamics over time. This memory advantage can be particularly beneficial when working with large text corpora or resource-constrained environments. We see this to be quite the advantage as memory intensive models will lead to a more exclusive research community. We detail some of the mathematics regarding the convergence and time complexity further in the appendix (Section \ref{time}, \ref{conv})

\section{Conclusion}

This work has emphasized the potential of Neural ODEs in processing textual data, particularly noting their ability to address the challenges of transparency in deep learning architectures. We hope future researchers will use this work to build more sophisticated architectures for interpretable text based classifiers even with the emergence of Large Language Model technologies since we only showcased this method at its simplest form.

\bibliography{colm2025_conference}
\bibliographystyle{colm2025_conference}

\newpage
\appendix

\section{Appendix}
\label{sec:appendix}

\subsection{Neural ODEs Architecture}
\label{arc}

Fundamentally a Neural ODE is easy to implement. The first block defines \texttt{ODEFunc}, a PyTorch module representing the derivative function of a Neural ODE, which applies a linear transformation followed by a ReLU activation to its input. The second block introduces \texttt{ODEBlock}, another PyTorch module that integrates the derivative function defined in \texttt{ODEFunc} over time, from \(t=0\) to \(t=1\), using the \texttt{odeint} solver to produce the system's state at \(t=1\).
The complexity of the model can be further extended for potential better performance, but the simplicity in this model also showed strong performance as seen in the study.
\begin{python}
# Define the ODE function (derivative)
class ODEFunc(nn.Module):
  def __init__(self, dim):
    super().__init__()
    self.linear = nn.Linear(dim, dim)
    self.relu = nn.ReLU(inplace=True)

  def forward(self, t, x):
    return self.relu(self.linear(x))

\end{python}

\begin{python}
# Define the ODE block that integrates 
# the ODEFunc
class ODEBlock(nn.Module):
    def __init__(self, odefunc):
        super().__init__()
        self.odefunc = odefunc

    def forward(self, x):
        # Integrate from t=0 to t=1 \
        return odeint(self.odefunc, x, \
        torch.tensor([0, 1],\
        dtype=torch.float32),\ 
        method='dopri5')[1]
\end{python}

\subsection{Understanding the Methodology through MNIST}

\begin{figure}[ht]
    \centering
    \includegraphics[width=0.8\columnwidth]{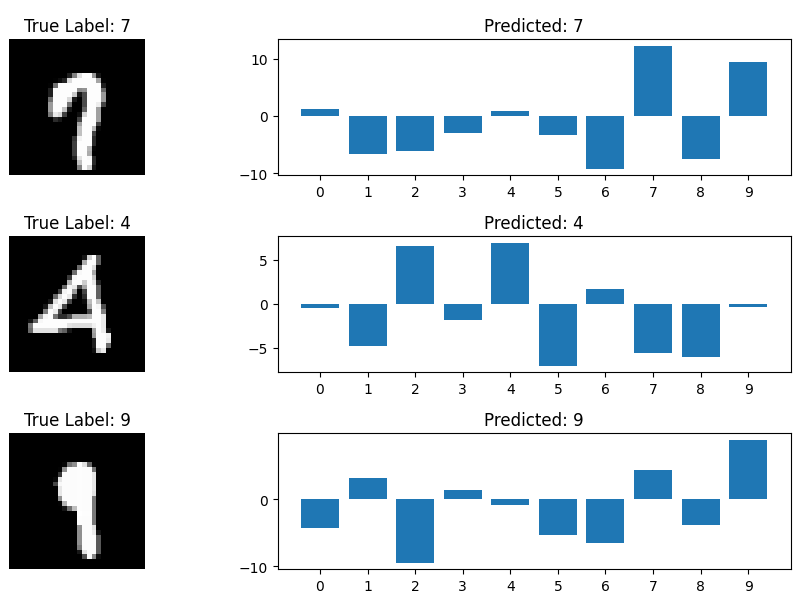}
    \caption{Showing the Probabilities of the Input Image on the MNIST Dataset during inference}
    \label{mnist}
\end{figure}

\paragraph{Neural Ordinary Differential Equations for MNIST Image Classification}

In this section, we explore the application of neural ordinary differential equations (neural ODEs) for the MNIST image classification task. The MNIST dataset consists of handwritten digits from 0 to 9, and the goal is to build a model that can accurately classify the given images into their respective digit classes.

Let $\mathcal{X} = \mathbb{R}^{28 \times 28}$ be the input space of MNIST images and $\mathcal{Y} = \{0, 1, \dots, 9\}$ be the set of digit class labels. Our objective is to learn a function $f: \mathcal{X} \rightarrow \mathcal{Y}$ that can predict the digit class of a given MNIST image.

We can formulate the MNIST image classification task as a neural ODE problem, where the dynamics of the image representation are modeled by an ODE:

$$\frac{\mathrm{d}x(t)}{\mathrm{d}t} = f(x(t), t, \theta)$$

where $x(t) \in \mathbb{R}^d$ is the hidden state representing the image at time $t$, $f: \mathbb{R}^d \times \mathbb{R} \times \mathbb{R}^p \rightarrow \mathbb{R}^d$ is the neural network-based function that governs the dynamics of the image representation, and $\theta \in \mathbb{R}^p$ are the parameters of the neural network.

To train the neural ODE model, we can use a dataset $\mathcal{D} = \{(x_i, y_i)\}_{i=1}^N$, where $x_i \in \mathcal{X}$ is the input MNIST image and $y_i \in \mathcal{Y}$ is the corresponding digit class label. We can define the following loss function:

$$\mathcal{L}(\theta) = \frac{1}{N} \sum_{i=1}^N \ell(f(x_i, \theta), y_i)$$

where $\ell: \mathcal{Y} \times \mathcal{Y} \rightarrow \mathbb{R}$ is the cross-entropy loss function.

To optimize the parameters $\theta$, we can use gradient-based methods, such as gradient descent or its variants. The gradients of the loss function with respect to the parameters $\theta$ can be computed using the adjoint method, which has a time complexity of $\mathcal{O}(k \cdot d)$, where $k$ is the number of stages in the numerical method used to solve the ODE.

Once the neural ODE model is trained, we can use it to predict the digit class of a new MNIST image $x$ by solving the ODE:

$$\frac{\mathrm{d}x(t)}{\mathrm{d}t} = f(x(t), t, \theta^*)$$

where $\theta^*$ are the learned parameters. The final digit class prediction can be obtained by applying a softmax layer to the terminal state $x(T)$, where $T$ is the final time point.

The key advantage of using a neural ODE for MNIST image classification is its ability to capture the dynamic evolution of the image representation, which can be particularly useful for modeling the spatial and temporal dependencies in the image data. Moreover, the continuous-time nature of the ODE allows for more flexible and expressive representations compared to traditional discrete-time models.

\subsection{Neural ODEs: Overcoming the Exploding Gradients Dilemma in Text Classification}
\label{grad}

One of the key advantages of neural ordinary differential equations for text classification is their ability to mitigate the problem of exploding gradients, a common issue that plagues many traditional deep learning architectures. Exploding gradients occur when the gradients of the loss function with respect to the model parameters grow exponentially during the training process, leading to numerical instability and poor model performance.

The inherent structure of neural ODEs can be mathematically formulated as follows. Consider a neural ODE of the form:

$$
\frac{\mathrm{d}x(t)}{\mathrm{d}t} = f(x(t), t, \theta)
$$

where $x(t)$ is the state of the system at time $t$, $f$ is the neural network-based function that governs the dynamics of the system, and $\theta$ is the set of parameters of the neural network.

The key property that allows neural ODEs to overcome the exploding gradients problem is the smoothness and stability of the ODE formulation. Unlike traditional recurrent neural networks (RNNs) or deep feedforward networks, which can suffer from vanishing or exploding gradients due to the compounding of operations across many layers, neural ODEs leverage the stability and convergence properties of ordinary differential equations.

Mathematically, the gradients of the neural ODE solution $x(t)$ with respect to the parameters $\theta$ can be computed using the adjoint method, which involves solving an additional ODE system:

$$
\frac{\mathrm{d}a(t)}{\mathrm{d}t} = -a(t)^\top \frac{\partial f}{\partial x}(x(t), t, \theta)
$$

where $a(t)$ is the adjoint state.

The key difference is that the gradients in the neural ODE formulation do not grow exponentially, but rather evolve smoothly over time, preserving the stability of the training process. This is due to the fact that the gradients are governed by the continuous-time dynamics of the ODE system, rather than being compounded across discrete layers.

\subsection{Convergence Proof for Training Neural Ordinary Differential Equations}
\label{time}

When training neural networks to model dynamical systems it is crucial to establish theoretical guarantees on the convergence of the training process. To achieve this, we provide a rigorous convergence proof for training neural ODEs.

The neural ODE is defined as $\frac{\mathrm{d}x(t)}{\mathrm{d}t} = f(x(t), t, \theta)$, where $x(t)$ is the state of the system, $f$ is the neural network-based function that governs the dynamics, and $\theta$ is the set of parameters. The training problem is formulated as an optimization problem, where the goal is to minimize the loss function $L(\theta) = \frac{1}{N} \sum_{i=1}^N \|x(t_i) - x_i\|^2$, where $x_i$ are the observed states at times $t_i$.

To prove the convergence, we make the following assumptions: (1) the neural network function $f$ is Lipschitz continuous with respect to $x$ and $\theta$, (2) the neural ODE has a unique solution $x(t)$ for any given initial condition and parameters, and (3) the optimization algorithm used to train the neural network converges to a stationary point of the loss function. Under these assumptions, we show that the neural ODE solution $x(t)$ will converge to the true state of the system as the number of training iterations goes to infinity.

\subsection{Time Complexity}
\label{conv}

The time complexity of training a neural ODE is analyzed in terms of the forward pass (solving the ODE to obtain $x(t_i)$), the backward pass (computing the gradients), and the optimization algorithm. The overall time complexity is $\mathcal{O}(N \cdot k \cdot n / \epsilon)$, where $N$ is the number of data points, $k$ is the number of stages in the numerical method used to solve the ODE, $n$ is the number of state variables, and $\epsilon$ is the desired accuracy of the optimization algorithm. The key factors that influence the time complexity are the number of data points, the complexity of the ODE solver, and the convergence rate of the optimization algorithm.

\subsection{Call for more sophisticated Neural ODE models}

In this work, we showcased a simple demonstration of how these models are interpretable and the advantages to modeling text continuously. Our NODE model, while not very sophisticated, demonstrates the potential of this approach. We see many directions for improvement, such as trying different ODE solvers, that could further optimize this model and make it an even better predictor.

We therefore encourage researchers to build upon this foundational NODE model and explore various avenues to enhance its performance. Some potential directions include experimenting with alternative ODE solvers, which may lead to improved stability, accuracy, and computational efficiency. Anther potential direction is leveraging auxiliary objectives, such as language modeling or semantic understanding, can help the NODE model learn more robust and transferable representations.

The versatility and dynamic modeling capabilities of neural ODEs make them a promising avenue for pushing the direction of interpretable text classification and unlocking new research in natural language processing.

\subsection{Sentiment Analysis with Neural Ordinary Differential Equations}

Sentiment analysis is a fundamental task in natural language processing, where the goal is to classify a given text as expressing a positive, negative, or neutral sentiment. In this section of the appendix, we present a theoretical framework for sentiment analysis using neural ordinary differential equations (neural ODEs) to help bridge the gap between text classification and NODEs.

Let $\mathcal{X}$ be the input space of text documents and $\mathcal{Y} = \{-1, 0, 1\}$ be the set of sentiment labels, where $-1$ represents negative sentiment, $0$ represents neutral sentiment, and $1$ represents positive sentiment. Our objective is to learn a function $f: \mathcal{X} \rightarrow \mathcal{Y}$ that can accurately predict the sentiment of a given text.

We can formulate the sentiment analysis task as a neural ODE problem, where the dynamics of the text representation are modeled by an ODE:

$$\frac{\mathrm{d}x(t)}{\mathrm{d}t} = f(x(t), t, \theta)$$

where $x(t) \in \mathbb{R}^d$ is the hidden state representing the text at time $t$, $f: \mathbb{R}^d \times \mathbb{R} \times \mathbb{R}^p \rightarrow \mathbb{R}^d$ is the neural network-based function that governs the dynamics of the text representation, and $\theta \in \mathbb{R}^p$ are the parameters of the neural network.

To train the neural ODE model, we can use a dataset $\mathcal{D} = \{(x_i, y_i)\}_{i=1}^N$, where $x_i \in \mathcal{X}$ is the input text and $y_i \in \mathcal{Y}$ is the corresponding sentiment label. We can define the following loss function:

$$\mathcal{L}(\theta) = \frac{1}{N} \sum_{i=1}^N \ell(f(x_i, \theta), y_i)$$

where $\ell: \mathcal{Y} \times \mathcal{Y} \rightarrow \mathbb{R}$ is a suitable loss function, such as the cross-entropy loss.

To optimize the parameters $\theta$, we can use gradient-based methods, such as gradient descent or its variants. The gradients of the loss function with respect to the parameters $\theta$ can be computed using the adjoint method, which has a time complexity of $\mathcal{O}(k \cdot d)$, where $k$ is the number of stages in the numerical method used to solve the ODE.

Once the neural ODE model is trained, we can use it to predict the sentiment of a new text $x$ by solving the ODE:

$$\frac{\mathrm{d}x(t)}{\mathrm{d}t} = f(x(t), t, \theta^*)$$

where $\theta^*$ are the learned parameters. The final sentiment prediction can be obtained by applying a classifier to the terminal state $x(T)$, where $T$ is the final time point.

The key advantage of using a neural ODE for sentiment analysis is its ability to capture the dynamic evolution of the text representation, which can be particularly useful for modeling the temporal and contextual aspects of language. Moreover, the continuous-time nature of the ODE allows for more flexible and expressive representations compared to traditional discrete-time models.

\begin{table*}
  \centering
  \begin{tabular}{llllll}
    \hline
    \textbf{Model} & \textbf{Interpretable?} & \textbf{Accuracy} & \textbf{F1} & \textbf{AUROC} & \textbf{}\\
    \hline
    Logistic Regression &  yes & 0.655 & 0.654 & 0.702\\
    LGBM     &  yes& 0.656 & 0.655 & 0.818  \\
    LSTM & no & 0.671 &  0.672 & 0.753 \\
    BERT & no & 0.737 & 0.738 & 0.799 \\
    Neural ODE &  yes& 0.731 & 0.721& 0.7209\\
    
    \hline
  \end{tabular}
  \caption{Benchmark of text classification (Discharge Location). Prevalence of this label was approximately 45\%.}
  \label{tab11}
\end{table*}

\begin{table*}
  \centering
  \begin{tabular}{llllll}
    \hline
    \textbf{Model} & \textbf{Interpretable?} & \textbf{Accuracy} & \textbf{F1} & \textbf{AUROC} & \textbf{}\\
    \hline
    Logistic Regression &  yes & 0.971 & 0.095 & 0.768\\
    LGBM     &  yes& 0.973 & 0.127 & 0.815  \\
    LSTM & no & 0.965 &  0.067 & 0.512 \\
    BERT & no & 0.988 & 0.072 & 0.546 \\
    Neural ODE &  yes& 0.970 & 0.069& 0.529\\
    
    \hline
  \end{tabular}
  \caption{Benchmark of text classification (Predicting Mortality). Prevalence of this label was approximately 3\%.}
  \label{tab14}
\end{table*}

\begin{table*}
  \centering
  \begin{tabular}{llllll}
    \hline
    \textbf{Model} & \textbf{Interpretable?} & \textbf{Accuracy} & \textbf{F1} & \textbf{AUROC} & \textbf{}\\
    \hline
    Logistic Regression &  yes & 0.835 & 0.427 & 0.807\\
    LGBM     &  yes& 0.877 & 0.545 & 0.818  \\
    LSTM & no & 0.881 &  0.543 & 0.799 \\
    BERT & no & 0.898 & 0.572 & 0.870 \\
    Neural ODE &  yes& 0.880 & 0.552 & 0.728\\
    
    \hline
  \end{tabular}
  \caption{Benchmark of text classification. (ICU Requirement) Prevalence of this label was approximately 19\%.}
  \label{tab13}
\end{table*}

\end{document}